\documentclass[conference]{IEEEtran}
\IEEEoverridecommandlockouts

\usepackage{graphicx}
\usepackage{amsmath}

\usepackage{amssymb}
\usepackage[english]{babel}
\usepackage{amsthm}
\usepackage{booktabs}
\usepackage{float}
\usepackage{color}
\usepackage[linesnumbered,ruled,vlined]{algorithm2e}
\usepackage{algorithmic}
\usepackage{diagbox}
\usepackage{multirow}
\usepackage{booktabs}
\usepackage{cuted}

\usepackage[capitalize]{cleveref}
\crefname{section}{Sec.}{Secs.}
\Crefname{section}{Section}{Sections}
\Crefname{table}{Table}{Tables}
\crefname{table}{Tab.}{Tabs.}

\usepackage{bm}

\usepackage{newfloat}
\usepackage{listings}

\usepackage{cite}
\usepackage{amsmath,amssymb,amsfonts}
\usepackage{algorithmic}
\usepackage{graphicx}
\usepackage{capt-of}
\usepackage{textcomp}
\usepackage{xcolor}
\def\BibTeX{{\rm B\kern-.05em{\sc i\kern-.025em b}\kern-.08em
    T\kern-.1667em\lower.7ex\hbox{E}\kern-.125emX}}
\begin{document}

\title{TD-BFR: Truncated Diffusion Model for Efficient Blind Face Restoration
\thanks{
\textsuperscript{†} Corresponding author.}
}

\author{
\IEEEauthorblockN{Ziying Zhang\textsuperscript{1} 
\hspace{1mm} Xiang Gao\textsuperscript{1}
\hspace{1mm} Zhixin Wang\textsuperscript{1}
\hspace{1mm} Qiang Hu\textsuperscript{1†}
\hspace{1mm} Xiaoyun Zhang\textsuperscript{1†}
} 
\IEEEauthorblockA{\textsuperscript{1}Cooperative Medianet Innovation Center, Shanghai Jiao Tong University, Shanghai, China\\
\textit{\{zyzhang2000, frxg0918, dedsec\_z, qiang.hu, xiaoyun.zhang\}@sjtu.edu.cn}}
\vspace{-10mm}
}

\maketitle

\begin{abstract}
Diffusion-based methodologies have shown significant potential in blind face restoration (BFR), leveraging their robust generative capabilities. However, they are often criticized for two significant problems: 1) slow training and inference speed, and 2) inadequate recovery of fine-grained facial details. To address these problems, we propose a novel Truncated Diffusion model for efficient Blind Face Restoration (TD-BFR), a three-stage paradigm tailored for the progressive resolution of degraded images. Specifically, TD-BFR utilizes an innovative truncated sampling method, starting from low-quality (LQ) images at low resolution to enhance sampling speed, and then introduces an adaptive degradation removal module to handle unknown degradations and connect the generation processes across different resolutions. Additionally, we further adapt the priors of pre-trained diffusion models to recover rich facial details. Our method efficiently restores high-quality images in a coarse-to-fine manner and experimental results demonstrate that TD-BFR is, on average, \textbf{4.75$\times$} faster than current state-of-the-art diffusion-based BFR methods while maintaining competitive quality.
\end{abstract}


\vspace{-5mm}
\section{Introduction}
\label{sec:intro}

Image restoration aims to reconstruct a high-quality image from its LQ observation. In typical image restoration problems such as denoising, deblurring, and super-resolution, the degradation process is simple and known (e.g., bicubic downsampling), while blind face restoration (BFR) endeavors to recover high-quality facial images from their degraded counterparts, which may involve various forms of degradation.

Recent research efforts like CodeFormer~\cite{CF}, GFPGAN~\cite{GFPGAN} integrate potent generative facial priors, to produce faithful and high-quality facial details. In recent years, denoising diffusion probabilistic models~\cite{ILVR14} have demonstrated remarkable performance in image generation. Building upon this, various approaches, such as DR2~\cite{dr2} and StableSR~\cite{stablesr}, have integrated DDPMs as an additional prior, enhancing their generative capabilities beyond traditional GAN-based methods. However, these methods prioritize image quality, sacrificing time efficiency by relying on complete diffusion model processes. Specifically, while StableSR and DifFace~\cite{difface} have demonstrated promising results with the help of powerful priors of diffusion models, it demands more than 50 sampling steps for restoring a single LQ face image, resulting in significant time expenditure. 
 To improve sampling efficiency in diffusion models, TDPM~\cite{TDPM} and RDM~\cite{relaydiffusion} propose to synthesize the image by merely using the partial generation process, which, in a faster manner, adds noise not until the data becomes pure random noise and use fewer reverse steps to generate data. Meanwhile, for the BFR task, EDBFR~\cite{EDBFR} addresses the lengthy sampling steps of diffusion-based methods by designing additional network modules to accelerate. However, the restored images often suffer from blurriness lacking details. Consequently, there is a critical demand for developing more efficient methodologies capable of ensuring high-quality face restoration images without compromising sampling efficiency.

\begin{figure}[t]
    \centering
    \vspace{-5mm}
    \includegraphics[width=1.0\columnwidth]{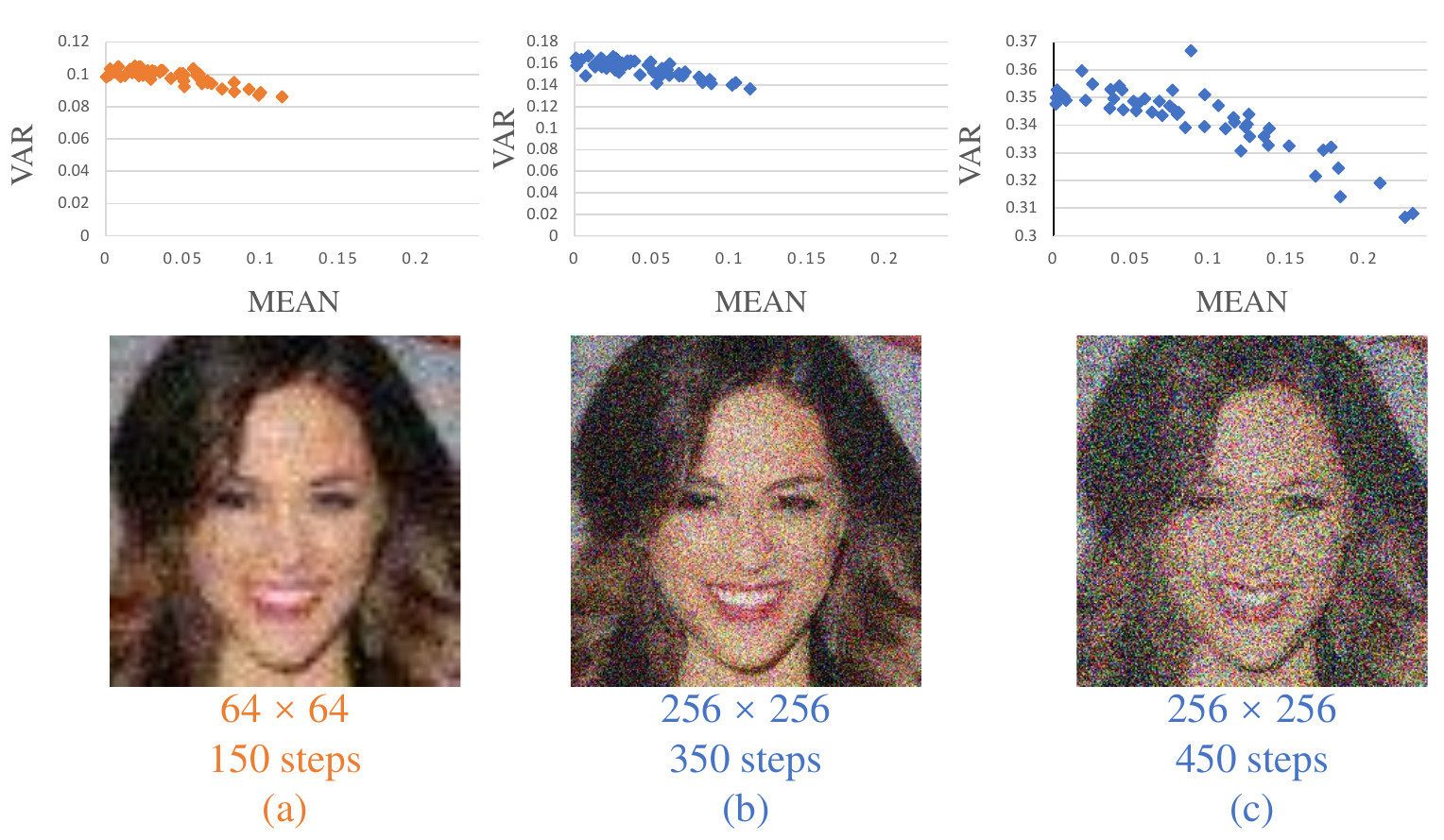}
    \vspace{-7mm}
    \caption{(a) presents the mean and variance of $(q(x_t|x_0) - q(x_0))$, and the visualized result at 150 steps for the 64-resolution diffusion model. (b) and (c) shows the corresponding data at 350 and 450 steps for the 256 resolution.}
    \vspace{-2mm}
    \label{Fig:distribu}
    \vspace{-2mm}
\end{figure}

In this paper, we propose the \textbf{T}runcated \textbf{D}iffusion model for \textbf{BFR} (\textbf{TD-BFR}), which adopts a three-stage paradigm designed to enhance the efficiency of diffusion-based methods while preserving the restoration quality. Inspired by ~\cite{relaydiffusion}, we find that the intermediate variables of diffusion models at different resolutions exhibit similar noise distributions across specific steps (shown in (a) and (b), distinct from (c)  of \cref{Fig:distribu}), mitigating the effects of degradation and resolution discrepancies. Therefore, our method decomposes the BFR task into a series of multi-stage processing steps to explicitly reduce the time burden of the restoration process by progressively addressing different resolutions of the input images. Initially, we employ a \textit{Low-resolution Startup (LRS)} to integrate low-quality images into the diffusion model by exchanging their low-frequency components with those of the intermediate variable using a low-pass filter at low resolution. Next, the \textit{Adaptive Degradation Remover (ADR)} dynamically selects the cutoff frequency to produce a smooth face image with reduced fine-grained details. Finally, we use \textit{Generative Detail Boost (GDB)}, which fine-tunes pre-trained text-to-image diffusion models~\cite{ldm} to generate accurate and visually pleasing image details. Quantitative experiments demonstrate that TD-BFR significantly improves sampling efficiency, achieving a restoration speed of \textbf{3.53s} per image while  other state-of-the-art diffusion based methods need  8.76s $\sim$ 25.05s (\cref{Tab:efficiency}). The main contributions of our paper are as follows:
\begin{itemize}
    \item [$\bullet$] We propose TD-BFR, a method that leverages the partial generation process to decompose the BFR task into three progressive multi-scale stages, significantly enhancing the sampling speed.
    \item [$\bullet$] We propose ADR, a module that dynamically addresses unknown degradation while serving as a bridge between the low-resolution integration in the first stage and the detail enhancement in the third stage, ensuring high-quality face restoration.
    \item [$\bullet$] Comprehensive experiments demonstrate the superiority of methods in both competitive quality and efficiency with sampling speed enhancement of \textbf{4.75$\times$} on average.
 
\end{itemize}

\vspace{-2mm}
\section{Methodology}
\vspace{-1mm}
\subsection{Preliminary}
Denoising Diffusion Probabilistic Models (DDPM) ~\cite{ILVR14} are a class of generative models that first pre-defines a variance schedule $\{\beta_1, \beta_2,...,\beta_T\}$ to progressively corrupt an image $x_0$ to a noisy status through forward (diffusion) process:
\begin{equation}
    q(x_t | _{t-1}) = \mathcal{N}(x_t; \sqrt{1 - \beta_t}x_{t-1}, \beta_t \mathbf{I}),
    \label{Eq:forward}
\end{equation}
Moreover, based on the property of the Markov chain, for any intermediate timestep $t\in\{1,2,..., T\}$, the corresponding noisy distribution has an analytic form:
\begin{align}
    q(x_t|x_0) &= \mathcal{N}(x_t; \sqrt{\bar \alpha_t}x_{0},   (1 - \bar \alpha_t)\mathbf{I}) \notag \\
                     &= \sqrt{\bar \alpha_t} x_0 + \sqrt{1 - \bar \alpha_t} \epsilon,
    \label{Eq:forw_detail}
\end{align}
\noindent where $\bar{\alpha}_t := \prod_{s=1}^t (1 -\beta_s)$ and $\epsilon \sim \mathcal{N}(\mathbf{0}, \mathbf{I})$. Then $x_T \sim \mathcal{N}(\mathbf{0}, \mathbf{I})$ if $T$ is big enough, usually $T = 1000$.

The model progressively generates images by reversing the forward process. The generative process is also a Gaussian transition with the learned mean $\mu_\theta$:
\begin{equation}
    p_\theta(x_{t-1} | x_t) = \mathcal{N}(x_{t-1}; \mu_\theta(x_t, t), \sigma_t^2 \mathbf{I}),
    \label{Eq:reverse}
\end{equation}
\noindent where $\sigma_t$ is usually a pre-defined constant related to the variance schedule, and $\mu_\theta(x_t, t)$, is usually parameterized by a denoising U-Net $\epsilon_\theta(x_t, t)$ \cite{ILVR14} with the following equivalence:
\begin{equation}
    \mu_\theta(x_t, t) = \frac{1}{\sqrt{\alpha_t}}(x_t - \frac{1 - \alpha_t}{\sqrt{1 - \bar{\alpha}_t}} \epsilon_\theta(x_t, t)).
    \label{Eq:rev_detail}
\end{equation}

\subsection{Framework Overview}

In \cref{Fig:distribu}, we examine the noise level similarity of intermediate variables across resolutions and steps. The mean and variance of $(q(x_t|x_0) - q(x_0))$ are calculated to evaluate noise consistency, as shown in the first row of \cref{Fig:distribu}. The VAR-MEAN figures indicate that the noise level of the 150-step variable at 64 resolution aligns closely with the 350-step variable at 256 resolution but differs from the 450-step. These findings, consistent with the visual results in the second row of \cref{Fig:distribu}, demonstrate that smooth truncation-stitching can be achieved by adjusting breakpoints, enabling a progressive multi-stage, multi-resolution framework.

\begin{figure}[t]
    \centering
    \vspace{-12mm}
    \includegraphics[width=1.0\columnwidth, trim=10 15 0 0]{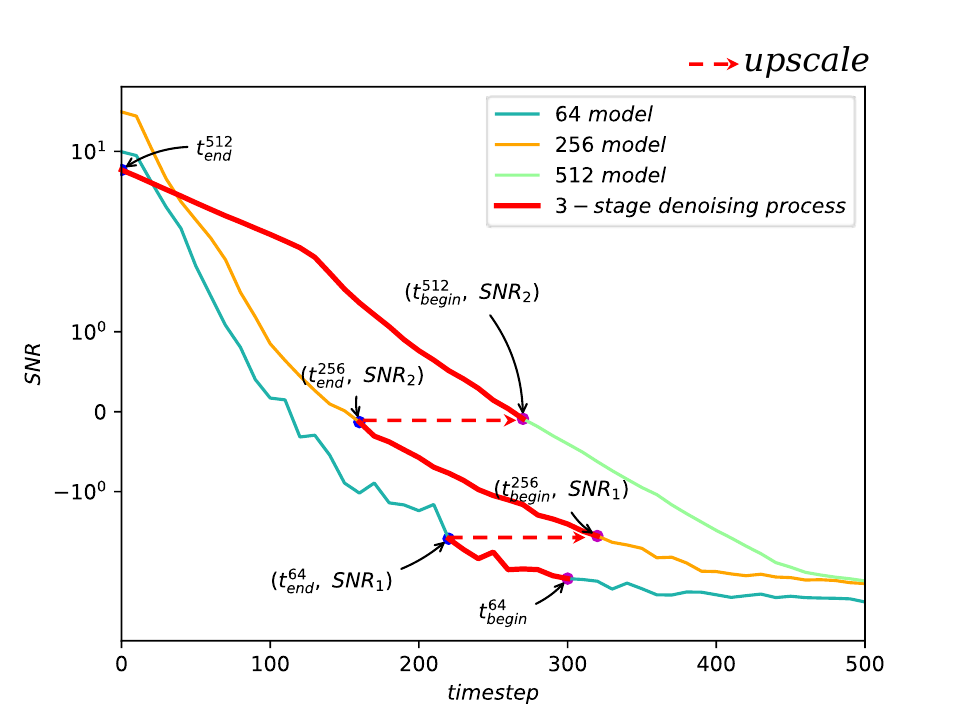}
    \caption{The Signal-to-noise ratio (SNR) curves of diffusion models at different resolutions. The breakpoints for truncated diffusion models can be selected based on similar SNR values, such as $SNR(t^{64}_{end}) \approx SNR(t^{256}_{begin})$.}
    
    \label{Fig:SNR}
    \vspace{-3mm}
\end{figure}

To better optimize breakpoint selection and model truncation across different resolutions, inspired by ~\cite{relaydiffusion}, we analyze the signal-to-noise ratio (SNR) curves of diffusion models at various resolutions, as shown in \cref{Fig:SNR}. The SNR quantifies the proportion of image semantics at the pixel level, and is defined by:
\begin{equation}
    SNR(t) := 10lg \frac{\sum^{M-1}_{i=0}\sum^{N-1}_{j=0}x_t(i,j)^2}{\sum^{M-1}_{i=0}\sum^{N-1}_{j=0} \left[x_t(i,j)-x_0(i,j)\right]^2}. \
    \label{SNR}
\end{equation}

Since SNR varies with resolution for the same number of steps, selecting appropriate breakpoints is essential for partial generative processes. To ensure consistent semantic proportions, breakpoints are determined based on comparable SNR values. For instance, transitioning from 64 to 256 resolution uses the $SNR_1$ value at the 64-resolution breakpoint as the starting point for the 256-resolution model, guiding step selection according to the lower-resolution SNR.

\begin{figure*}[t]
\vspace{-5mm}
    \centering
    \includegraphics[width=1.0\linewidth]{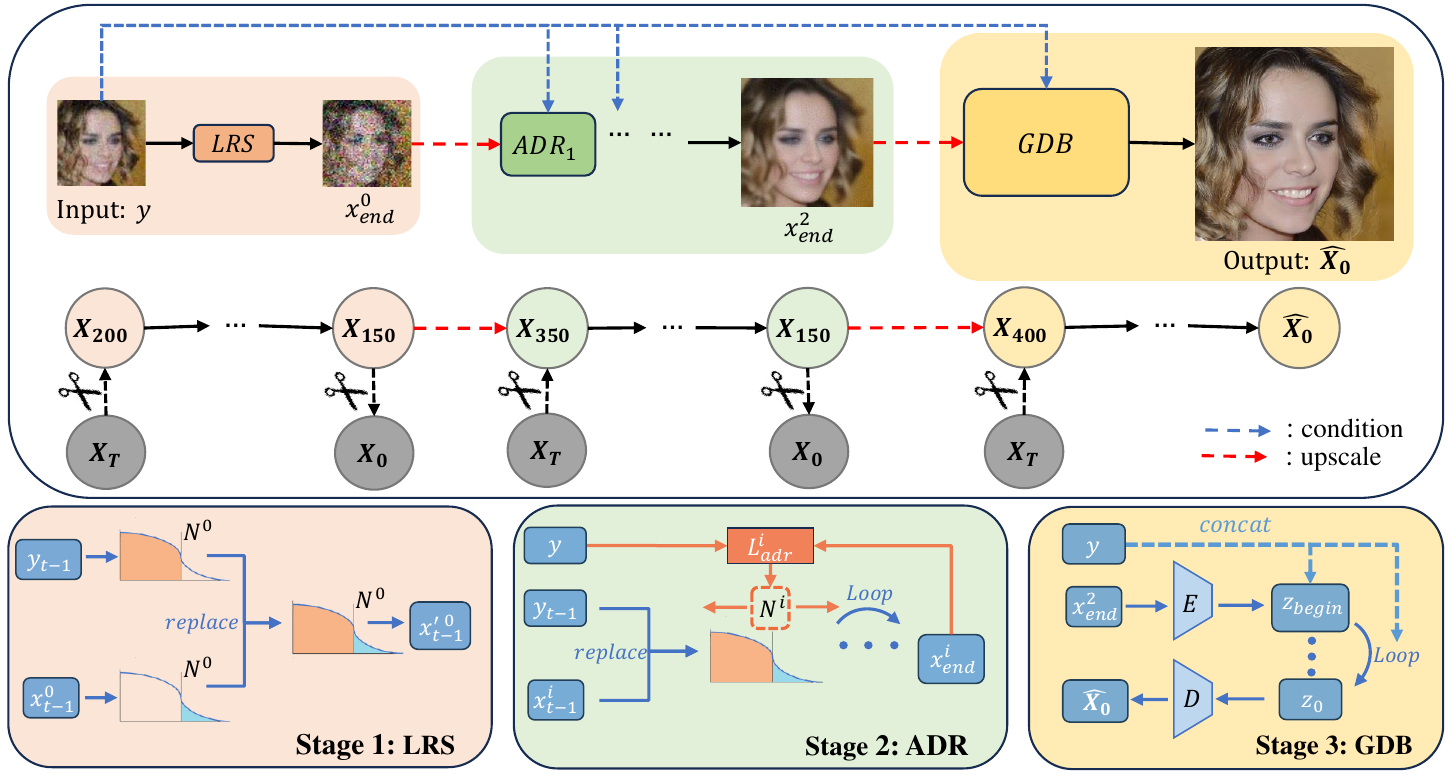}
    \caption{Overall TD-BFR: A three-stage paradigm tailoring for progressive resolution of the input image. ADR serves as a bridge for adaptive handling of unknown degradation and resolution variance, enabling low-resolution integration via LRS and detail generation via GDB.}
    \label{Fig:framework}
    \vspace{-4mm}
\end{figure*}
Building on this, our proposed truncated framework, TD-BFR, shown in \cref{Fig:framework}, adopts a three-stage paradigm to restore degraded images of varying resolutions progressively. The first stage, Low-Resolution Startup (LRS), integrates the LQ image into the diffusion model at a low resolution. The second stage, Adaptive Degradation Remover (ADR), addresses unknown degradations and transitions between resolutions. Finally, the third stage, Generative Detail Boost (GDB), leverages the fine-tuned diffusion model's priors to generate detailed, high-quality facial features, significantly enhancing restoration quality and sampling efficiency.

\begin{figure*}[t]
    \centering
    \includegraphics[width=1.0\textwidth]{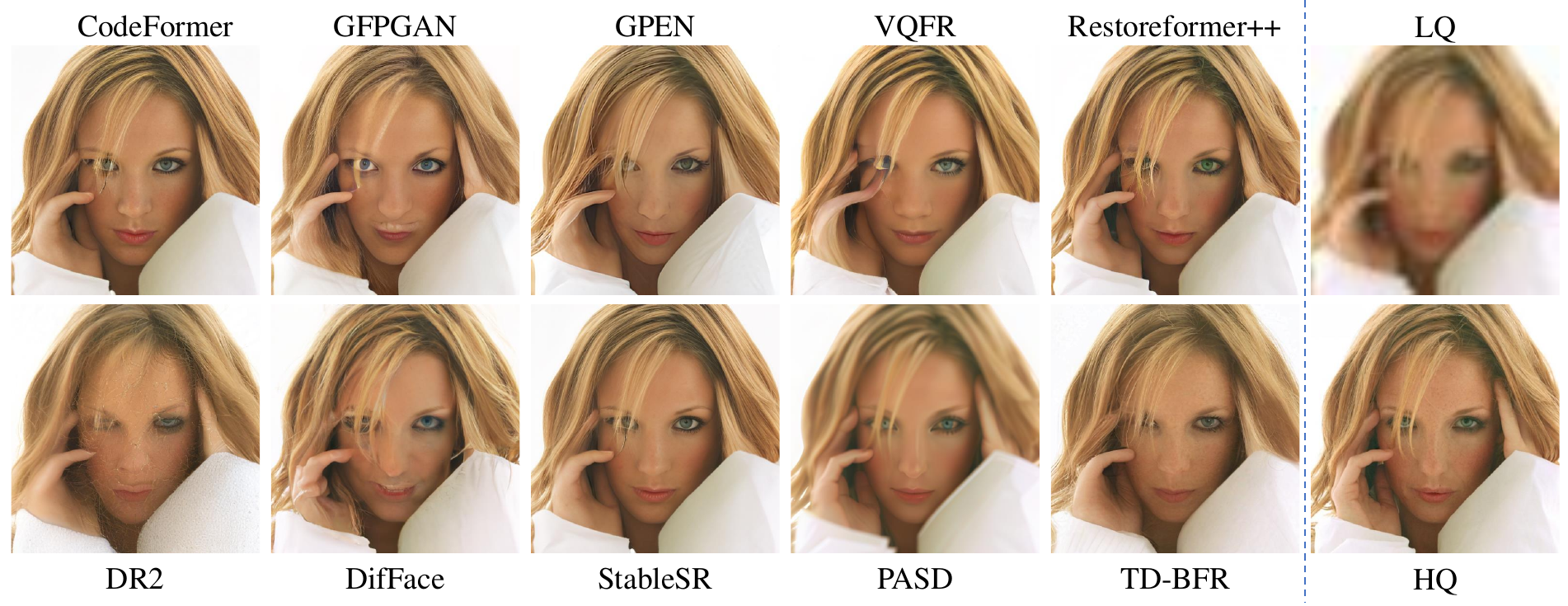}
    \vspace{-2mm}
    \caption{Qualitative comparisons on synthetic CelebA datasets. Our method (TD-BFR) achieves higher restoration quality, encompassing more precise facial details with fewer semantic alterations. (Zoom in for the best view)}
    \label{Fig:CelebaTest}
    
\end{figure*}

\subsection{Truncated Diffusion model for efficient BFR}
\textbf{Stage One: Low-resolution Startup(LRS).}
    Due to the substantial computational demands of high-resolution diffusion models, we advocate for implementing an LRS, such as a 64-resolution initialization, utilizing the input LQ $y$. Proved by ~\cite{dr2}, (1) there exists an intermediate timestep $t_{end}$ such that for $t > t_{end}$, the distance between $q(x_t|x)$ and $q(y_t|y)$ is close especially in the low-frequency part; (2) there exists $t_{begin} > t_{end}$ such that the distance between ($x_{t_{end}}$ is abbreviated as $x_{end}$) $q(x_{{begin}}|x)$ and $q(y_{{begin}}|y)$ is eventually small enough, satisfying ${q(x_{{begin}}|x)} \approx {q(y_{{begin}}|y)}$. 
    Hence, the initial stage involves employing a truncated diffusion process at lower resolutions, utilizing incomplete steps $\{t_{end}^0,..., t_{begin}^0\}$, thereby incorporating the LQ images into the entirety of the progressive multi-stage framework.

Let $\Phi_{N^0}(\cdot)$ denote a linear low-pass filtering operation, a sequence of downsampling and upsampling by an initial factor of $N^0$, therefore maintaining the dimensionality of the image. After each transition from $x_t$ to $x_{t-1}$, we sample $y_{t-1}$ from $y$ through \cref{Eq:forw_detail}. Leveraging the linear property of $\Phi_{N^0}$, we substitute the low-frequency component of $x_{t-1}$ with that of $y_{t-1}$ due to their proximity in distribution \cite{dr2}. This substitution can be expressed as:
\begin{equation}
        x_{t-1}^{' 0} := \Phi_{N^0}(y_{t-1}) + (\mathbf{I} - \Phi_{N^0})(x^0_{t-1}).
        \label{Eq:ILVR}
\end{equation}
In the stage 1 of \cref{Fig:framework}, the LRS module excludes the high-frequency component of $y$ due to its limited information from degradation. The use of a truncated low-resolution diffusion model introduces minimal time overhead. For BFR with unknown degradation, adding Gaussian noise multiple times to $x_t$ effectively conceals the degradation \cite{dr2}, while low-frequency information retains semantic relevance via linear low-pass filtering $\Phi_{N^0}(\cdot)$. Truncating the forward-reverse diffusion process at low resolution in this stage seamlessly integrates the LQ image into the progressive resolution framework.

\textbf{Stage Two: Adaptive Degradation Remover(ADR).}
Based on the severity of degradation caused by the upscaled resolution and the original LQ, ADR is proposed relying on the outcomes $x_{end}^0$ obtained from LRS. This module automatically adjusts the parameter $N^i$ of the i-th low-pass filter, thereby facilitating BFR and acting as a bridge for resolution transitions. The ADR module can be utilized multiple times at mid-stage to enhance degradation removal. In this study, we demonstrate its application twice at a resolution of 256 as an illustrative example.
For the i-th ADR and the partial time steps $t\in\ \{t_{end}^i,...,t_{begin}^i\}$ based on the SNR curve in \cref{Fig:SNR}, we combine the low-frequency of $x$ with the high-frequency of $y$ to obtain the optimized intermediate variable:
\begin{equation}
        x_{t-1}^{' i} := \Phi_{N^i}(y_{t-1}) + (\mathbf{I} - \Phi_{N^i})(x^i_{t-1}),
\end{equation}
 where $x_{t-1}^i$ denotes the original intermediate variable of the i-th mid-resolution diffusion model. Given the typically unknown severity of degradation in LQ  for BFR, previous methods often necessitated manual selection of filter parameters $N$ to extract low-frequency information~\cite{dr2, ILVR}. To streamline this process while maintaining effective degradation removal, we introduce an ADR, which computes the Mean Squared Error (MSE) loss $L_{adr}^i$ between the restorated result $x_{{end}}^i$ and the input $y$ at the final step, optimizing the parameter $N$ accordingly. The expression for $L_{adr}^i$ is formulated as follows:
\begin{equation}
    L_{adr}^i := \frac{1}{mn}\sum^m_{j=1}\sum^n_{k=1}||\Phi_{N^i}(y)(j,k) - \Phi_{N^i}(x_{{end}}^i)(j,k)||^2.
\end{equation}

When $L_{adr}^i$ exceeds the threshold, which will be discussed in the appendix, the ADR adaptively selects the parameter $N^i$ that matches the severity of the input image $y$ until $L_{adr}^i$ falls below the threshold, to ensure that the semantics of $x_{end}^i$ remain consistent with $y$ during degradation removal and to prevent the restoration result from gradually deviating from the input image $y$. Additionally, through this operation, the resolution impact caused by multi-scale effects is also mitigated.

\textbf{Stage Three: Generative Detail Boost(GDB).}
After the  ADR, which retains fundamental semantics but loses high-frequency details, we introduce GDB to enhance image details. GDB leverages the strong priors of large-scale text-to-image diffusion models, such as the Stable Diffusion Model (SDM)~\cite{ldm}, to restore realistic details.

In pursuit of refined facial priors, inspired by ~\cite{instructpix2pix}, we fine-tune the SDM by concatenating synthetic degraded images $x_{deg}$ with the noisy latent variable $z_t$ at time $t$ and inputting this into the denoising U-Net $\epsilon_\theta(z_t, t)$. Our optimization objective is to minimize the latent diffusion loss: 
\begin{equation}
        L = E_{\varepsilon(x),x_{deg},\epsilon \sim \mathcal{N}(\mathbf{0}, \mathbf{I}),t} \left[||\epsilon - \epsilon_\theta(cat(z_t, x_{deg}), t)||^2_2\right],
        \label{Eq:SDM}
\end{equation}

Considering the SNR characteristics of diffusion models, we use the fine-tuned truncated SDM in the third stage of our sampling process. The intermediate output $x_{end}^2$ from the second stage is encoded by $\varepsilon(\cdot)$ into latent space, then forwarded to produce the truncated latent result $z_{begin}$, initiating GDB. Breakpoints for ADR and GDB are chosen based on the SNR properties of the respective resolution.
In stage 3 of \cref{Fig:framework}, we illustrate the truncated sampling process post-SDM fine-tuning. By utilizing SNR properties and the priors of the pre-trained SDM, we perform incomplete denoising to maintain restoration quality comparable to other diffusion methods, while significantly reducing time consumption.

\begin{figure*}[t]
    \centering
    \includegraphics[width=\textwidth]{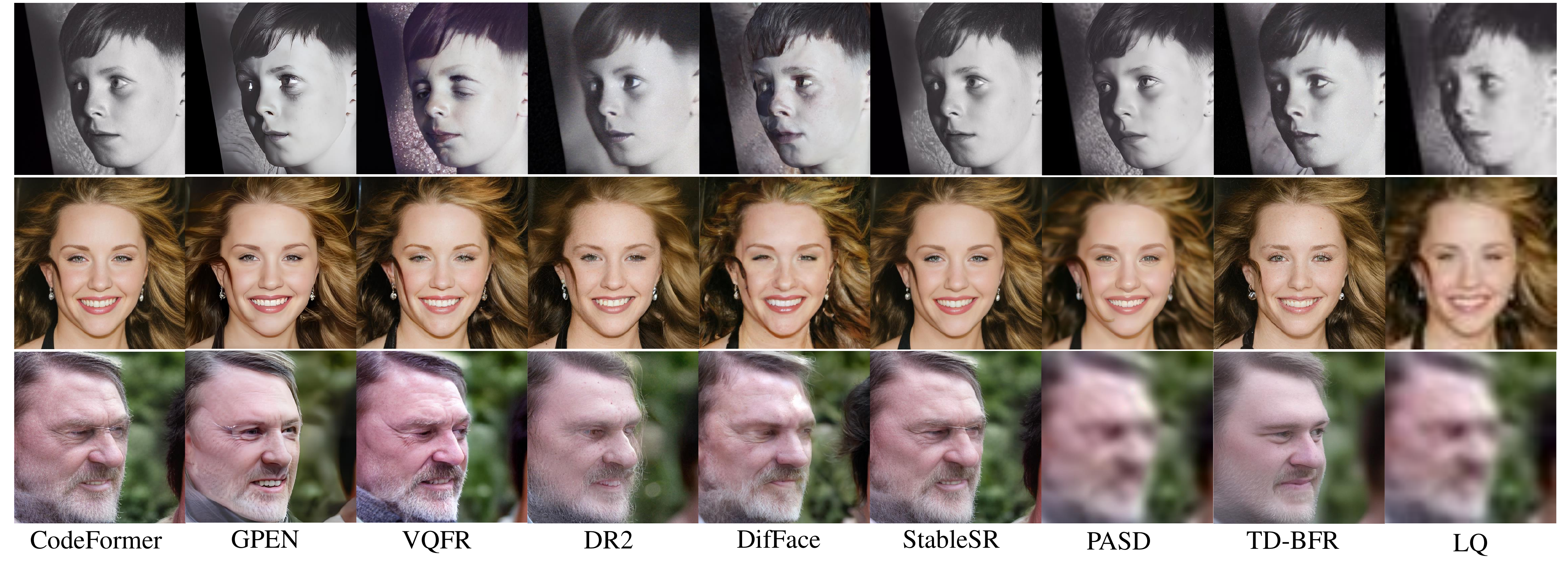}
    \vspace{-5mm}
    \caption{Qualitative comparison with SOTA BFR methods on Celeb-Child (first row), LFW-Test (second row), and WIDER-Test (third row).}
    \label{Fig:RealTest}
    
\end{figure*}

\begin{table*}[t]
  \caption{Efficiency comparison on CelebA-Test dataset. The Sampling Time Ratio (STR) compares the time consumption of other methods with ours, and shows that other methods are, on average, \textbf{4.75$\times$} slower than TD-BFR. \textbf{Bold} and \underline{underlined} indicate the best and the second best performance. }
  \centering
   {
      \begin{tabular}{c | c | c c c c | c c }
        \toprule
         Methods 
         & Params. (M) 
         & PSNR $\uparrow$ & LPIPS $\downarrow$ & NIQE $\downarrow$ & FID $\downarrow$ 
         & Time (s) & STR \\
        \midrule
        DR2 \cite{dr2}         &\textbf{93.56}      & 23.55 &0.434  &4.20   &50.13 &8.76  &2.48$\times$\\
        DifFace\cite{difface}   &\underline{175.38}     & 25.18 &0.461 & \underline{4.01}   &\underline{48.98}  &25.04 &7.09$\times$\\
        StableSR\cite{stablesr} &971.23    &24.99  &\underline{0.361}  &4.60   &50.50  &18.33  &5.19$\times$\\
        PASD\cite{pasd}         &1360.99       &\underline{25.19}  &0.419   &7.77   &67.05  &14.90 &4.22$\times$\\
        \midrule
        TD-BFR(Ours)                    &1229.48   &\textbf{25.21} 
        &\textbf{0.347} &\textbf{3.99}  &\textbf{43.92} &\textbf{3.53}   &1$\times$ \\
        \bottomrule
      \end{tabular}
    }
  
  \label{Tab:efficiency}
    \vspace{-3mm}
\end{table*}

\section{Experiments}

\subsection{Implementation, Datasets, Metrics}
\noindent \textbf{Implementation.} 
The complete diffusion models at 64 and 256 resolution employed by stages 1 \& 2 are trained on the FFHQ dataset~\cite{GFPGAN35}. Following this, we utilize Stable Diffusion 2.1-base~\cite{ldm} as the generative prior. More details on the training will be introduced in the supplementary material.

\noindent \textbf{Test Datasets}. We construct one synthetic dataset on 
CelebA-HQ~\cite{GFPGAN51} and three real-world datasets, including WIDER-Test~\cite{CF41}, CelebChild, and LFW-Test~\cite{GFPGAN29}  for testing. 
The specific explanation for datasets is detailed in
the supplementary.      

\noindent \textbf{Metrics.} 
For evaluation, we adopt two pixel-wise metrics (PSNR and SSIM), a reference perceptual metric (LPIPS), and two non-reference metrics (NIQE and FID). All metrics are used for the synthetic data while only NIQE and FID are used for real-world datasets due to the lack of HQ images. 

\vspace{-1mm}
\subsection{Efficiency comparison with diffusion-based methods}

Since TD-BFR aims to enhance the efficiency of diffusion-based BFR methods, we compare it exclusively with diffusion-based methods. We compare it with DR2~\cite{dr2} (using SparNetHD~\cite{SPAR} as its enhancement module), DifFace~\cite{difface}, StableSR~\cite{stablesr}, and PASD~\cite{pasd}, using their official codes and results. Efficiency is assessed via parameters and inference time, while quality is evaluated using PSNR, LPIPS, NIQE, and FID. As shown in \cref{Tab:efficiency}, TD-BFR outperforms in quality metrics and achieves an average speed improvement of \textbf{4.75$\times$}. Despite having more parameters, the truncated structure significantly reduces sampling steps, greatly enhancing efficiency.

\vspace{-1mm}
\subsection{Comparisons with State-of-the-Art Methods}

In our comparisons, we contrast TD-BFR with both diffusion-based methods in Efficiency Comparison, and non-diffusion-based methods, including Codeformer ~\cite{CF}, GFP-GAN ~\cite{GFPGAN}, GPEN~\cite{GPEN},  Restoreformer++~\cite{restoreformer++}, VQFR ~\cite{VQFR}, as well as methods based on diffusion models mentioned in Efficiency Comparison.

\noindent \textbf{Synthetic CelebA-Test}.
\cref{Tab:Celeba} presents quantitative comparisons on the synthetic CelebA-Test dataset. Our method excels in full-reference metrics (PSNR, LPIPS) and non-reference metric (NIQE), with SSIM and FID ranking second. \cref{Fig:CelebaTest} shows qualitative comparisons with SOTA methods. Non-diffusion methods often introduce artifacts, particularly around the mouth, while diffusion methods like DR2 and DifFace exhibit significant artifacts, and StableSR and PASD are overly smooth. In contrast, our method, utilizing diffusion priors and a truncated structure, generates high-quality facial images with minimal artifacts, preserving identity.

\begin{table}[t]
  \caption{Quantitative comparisons on CelebA-Test dataset. \textbf{Bold} and \underline{underlined} indicate the best and the second best performance.}
  \vspace{-2mm}
  \centering 
  
   \resizebox{\columnwidth}{23mm}{
      \begin{tabular}{c|c c c c c}
        \toprule
         Methods & PSNR $\uparrow$ & SSIM $\uparrow$ & LPIPS $\downarrow$ 
         & NIQE $\downarrow$ & FID $\downarrow$  \\
        \midrule
        CodeFormer        & 24.81 & 0.659  &\underline{0.349} &4.52  & 51.18   \\
         GFP-GAN      & 24.44 & 0.664  & 0.377    &4.28  & 48.19   \\
         GPEN           &24.26  &0.665  &0.371  &5.33   &49.66  \\
         VQFR           & 23.79 & 0.653  & 0.360    &4.17  & 46.07   \\
         Restoreformer++  &25.09  &0.657  &0.352  &4.12   &\textbf{39.98}\\
        
        \midrule
        DR2           & 23.55 & 0.595  & 0.434    &4.20  & 50.13   \\
        DifFace      &25.18 &0.680  &0.461  &\underline{4.01}   &48.98\\
        StableSR     & 24.99 & 0.671  & 0.361    &4.60  & 50.50   \\
        PASD       &\underline{25.19}  &\textbf{0.698} &0.419  &7.77   &67.05\\
        \midrule
         \textbf{TD-BFR(Ours)}      & \textbf{25.21} & \underline{0.683}  & \textbf{0.347}  & \textbf{3.99}    &\underline{43.92}   \\
        \bottomrule
      \end{tabular}
    }
  \vspace{-2mm}
  \label{Tab:Celeba}
  
\end{table}

\noindent \textbf{Real-world Datasets}.
\cref{Tab:real} shows quantitative comparisons of NIQE and FID on real-world datasets, where TD-BFR achieves the best NIQE and ranks second in FID. \cref{Fig:RealTest} presents three examples, excluding underperforming non-diffusion methods (GFPGAN and Restoreformer++) for brevity (full results in the supplementary material). Previous BFR methods often produce hazy or unnatural faces, especially under severe degradations (Row 3). In contrast, our method generates natural, detailed results, preserving features like wrinkles and eyes.

\begin{table}[t]
\caption{Quantitative comparisons of FID on LFW-Test, WIDER-Test, and CelebChild-Test. \textbf{Bold} and \underline{underlined} indicate the best and the second best performance.}
  \centering
  \resizebox{\columnwidth}{23mm}{
      \begin{tabular}{c | c c | c c | c c}
        \toprule
            & \multicolumn{2}{c}{\textbf{LFW-Test}} &\multicolumn{2}{c}{\textbf{ WIDER-Test}} & \multicolumn{2}{c}{\textbf{CelebChild-Test}}   \\
        \midrule
        Methods     & NIQE $\downarrow$ & FID $\downarrow$ 
                & NIQE $\downarrow$ & FID $\downarrow$
                & NIQE $\downarrow$ & FID $\downarrow$ \\
        \midrule
        CodeFormer       & 4.34 & 53.62 & 4.28 & 60.72 & 4.82 & 113.42 \\
         GFP-GAN      &4.36 & 53.19 &4.32   & 65.18 &4.65   & 114.19 \\
         GPEN       &5.28   &51.98  &5.38   &71.85  &5.57   &122.18 \\
         VQFR        &3.95  & 52.50 &4.10   &59.87 &4.50    & 112.90 \\
         Restoreformer++    &\underline{3.93}   &49.60  &\underline{3.93}   &\underline{58.27} &4.42   &\textbf{105.98}    \\
        
        \midrule
        DR2             &4.08   & 50.86 &4.32   & 67.44 &\underline{4.27}  & 121.53 \\
        DifFace         &3.98   &48.08  &4.22   &59.29  &4.58   &112.37 \\
        StableSR    &4.48   & 48.59 &4.38   & 58.32 &4.80   & 115.90 \\
        PASD        &5.52   &\textbf{46.35} &10.78  &139.96 &5.21   &135.53 \\
        \midrule
         \textbf{TD-BFR(Ours)}      &\textbf{3.91}  & \underline{47.88} &\textbf{3.88}  & \textbf{57.59} &\textbf{4.19}   & \underline{110.55} \\
        \bottomrule
      \end{tabular}
    }
  
  \label{Tab:real}
  \vspace{-2mm}
\end{table}

\vspace{-2mm}
\subsection{Ablation studies}
\begin{figure}
    \centering
    \includegraphics[width=1.0\columnwidth]{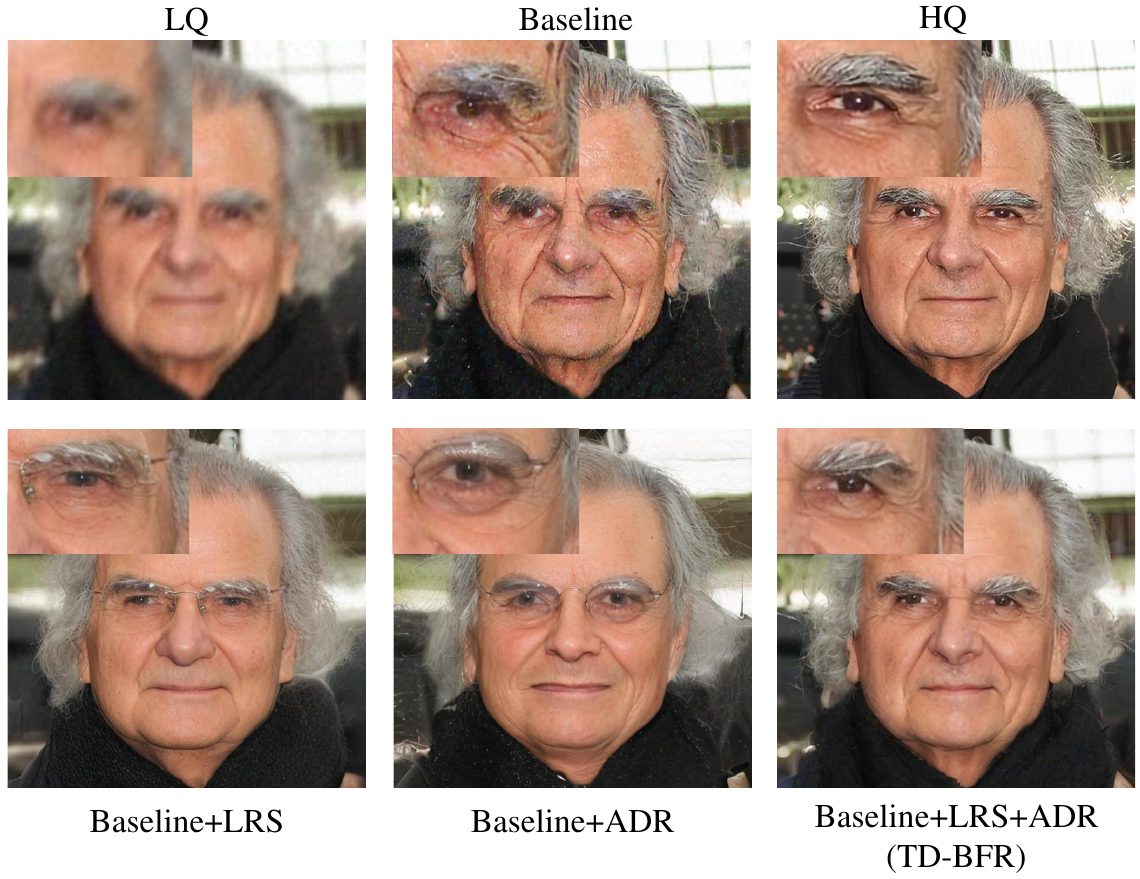}
     \vspace{-6mm}
    \caption{Visual comparison of ablation studies. (Zoom in for the best view)}
    \label{Fig:Ablation}
    \vspace{-3mm}
\end{figure}
\noindent \textbf{Baseline.} Since Stage 3, GDB serves as the detail generation module, it retains some restoration capabilities even without LRS and ADR. Therefore, we conducted ablation experiments using GDB as the baseline.

\noindent \textbf{Baseline + LRS}. LRS improves sampling efficiency but has weaker degradation removal capabilities, leading to GDB generating results lacking fine details (see \cref{Fig:Ablation}). The quantitative results are also significantly lower in \cref{Tab:ablation}.

\noindent \textbf{Baseline + ADR}. ADR, as the core degradation removal module, can produce comparable image quality when combined with the baseline (see \cref{Fig:Ablation}). However, the absence of LRS results in more truncation steps at higher resolution, significantly reducing efficiency (see \cref{Tab:ablation}).

\begin{table}[t]
 \caption{Ablation studies on CelebA-Test dataset. \textbf{Bold} indicates the best performance.}
  \centering
  \resizebox{\columnwidth}{12.5mm}{
      \begin{tabular}{c|c c c c | c}
        \toprule
        Methods & PSNR $\uparrow$ & SSIM $\uparrow$ 
        & LPIPS $\downarrow$ & FID $\downarrow$ 
        &Time(s)  \\
        \midrule
        Baseline & 24.21 & 0.642 & 0.435& 53.43 &5.27  \\
        Baseline+LRS         & 24.52 & 0.657 & 0.433 & 53.01 &4.68  \\
        Baseline+ADR         & 25.19 & 0.680 & 0.351 & 44.12 &7.07\\
        \midrule
         Baseline+LRS+ADR & \textbf{25.21} & \textbf{0.683} & \textbf{0.347}    &\textbf{43.92} &\textbf{3.53}   \\
         
        \bottomrule
      \end{tabular}
    }
 \vspace{-4mm}
  \label{Tab:ablation}
\end{table}
\vspace{-1mm}
\section{Conclusion}
We propose TD-BFR, a progressive framework for the BFR task, which significantly improves sampling speed by using incomplete diffusion models during inference. An ADR addresses unknown degradations and resolution variations, ensuring high-quality restoration through LRS and high-resolution GDB. Extensive experiments show that TD-BFR achieves comparable image quality to state-of-the-art methods while improving speed by \textbf{4.75$\times$} on average.

\vspace{-2mm}
\section{Acknowledgments}
This work is supported by National Natural Science Foundation of China (62271308),STCSM ( 24ZR1432000, 24511106902, 24511106900, 22511105700, 22DZ2229005), 111 plan (BP0719010), and State Key Laboratory of UHD Video and Audio Production and Presentation.



\bibliographystyle{IEEEbib}
\bibliography{main}

\end{document}


\title{TD-BFR: Truncated Diffusion Model for Efficient Blind Face Restoration supplementary material}

\author{Anonymous ICME submission}

\maketitle

\begin{strip}
\centering

\includegraphics[width=\textwidth]{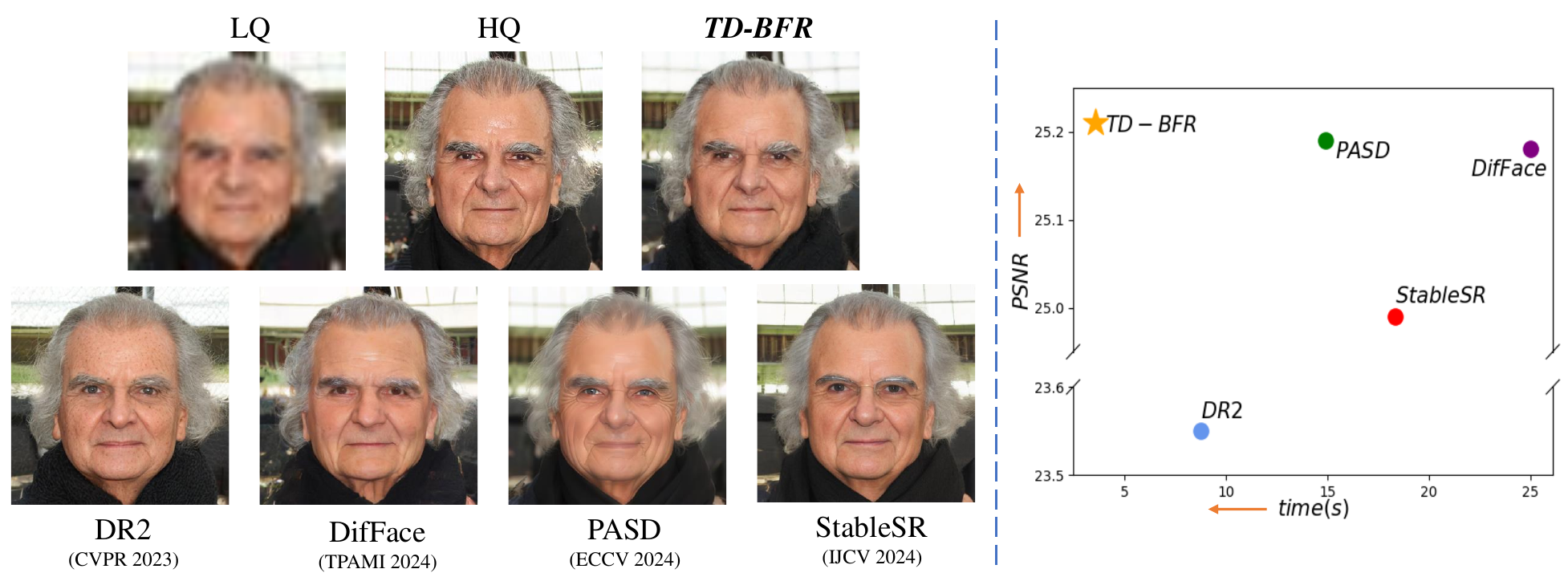}
  \captionof{figure}{
  \textbf{Left}: Compared to SOTA diffusion-based methods, TD-BFR restores LQ with richer details while remaining more faithful to the high-quality(HQ) images. \textbf{Right}: Compared to other state-of-the-art diffusion-based methods, our approach achieves higher PSNR while requiring much less sampling time.
  }
  \label{fig:teaser}

  \end{strip}

\begin{figure*}[htbp]
    \centering
    \vspace{-10mm}
    \includegraphics[width=\textwidth]{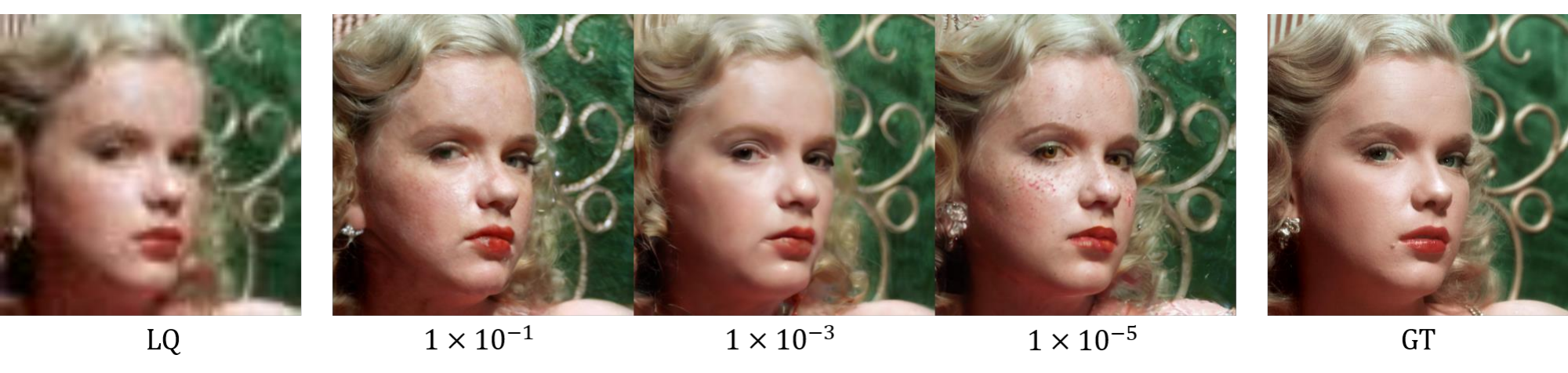}
    \caption{Visual differences caused by large-scale changes in threshold 1.}
    \label{Fig:Threshold1}
\end{figure*}

\section{Threshold Section}

\begin{table}[t]
\caption{Comparison of quantitative results for wide range threshold 1 changes. \textbf{Bold} indicates the best performance.}
    \centering
    \resizebox{\columnwidth}{!}{
    \begin{tabular}{c|c c c c c c}
    \toprule
         Threshold  & $1\times10^{-5}$  & $1\times10^{-4}$ & $1\times10^{-3}$ & $1\times10^{-2}$ & 0.1 & 1\\
         \midrule
         FID $\downarrow$ & 58.21 & 57.56 & \textbf{45.75}  & 46.35 & 58.58 & 57.94 \\
         \bottomrule
    \end{tabular}
    }
    \label{tab:threshold_wide}
\end{table}

\begin{table}[t]
\caption{Comparison of quantitative results for small range threshold 1 changes. \textbf{Bold} indicates the best performance.}
    \centering
    \resizebox{\columnwidth}{!}{
    \begin{tabular}{c|c c c c c c}
    \toprule
         Threshold & $1\times10^{-3}$ & $2\times10^{-3}$ & $4\times10^{-3}$ & $6\times10^{-3}$ & $8\times10^{-3}$ & $1\times10^{-2}$ \\
         \midrule
         FID  $\downarrow$ & 45.75  & \textbf{43.92} & 45.54& 46.92  & 47.90  &  46.35 \\
         \bottomrule
    \end{tabular}
    }
    
    \label{tab:threshold_small}
\end{table}

During the experiment, we find that the experimental results are not sensitive to the threshold within a certain order of magnitude. Still, once it exceeds the range, it may lead to unclean degradation removal or poor semantic information retention. Since each ADR module has equal status, we take threshold 1 as an example here. \cref{tab:threshold_wide} shows that when threshold 1 exceeds a too-large range, it will lead to poor results. As shown in \cref{tab:threshold_small}, when threshold 1 is within $1\times10^{-2}$ to $1\times10^{-3}$, there is no obvious difference in FID. As shown in the second column of \cref{Fig:Threshold1}, if it is larger than this range, It may lead to incomplete removal of degradation, which is reflected in the image result that the face is not smooth enough, while shown in the fourth column of \cref{Fig:Threshold1}, if smaller than this range, it may lead to severe loss of semantic information, ID Change, Such as rash on face and eye color.

\section{Low-Resolution Startup Initialization}

\begin{figure}[t]
    \centering
    \includegraphics[width = 1.0\columnwidth,trim=20 10 20 10]{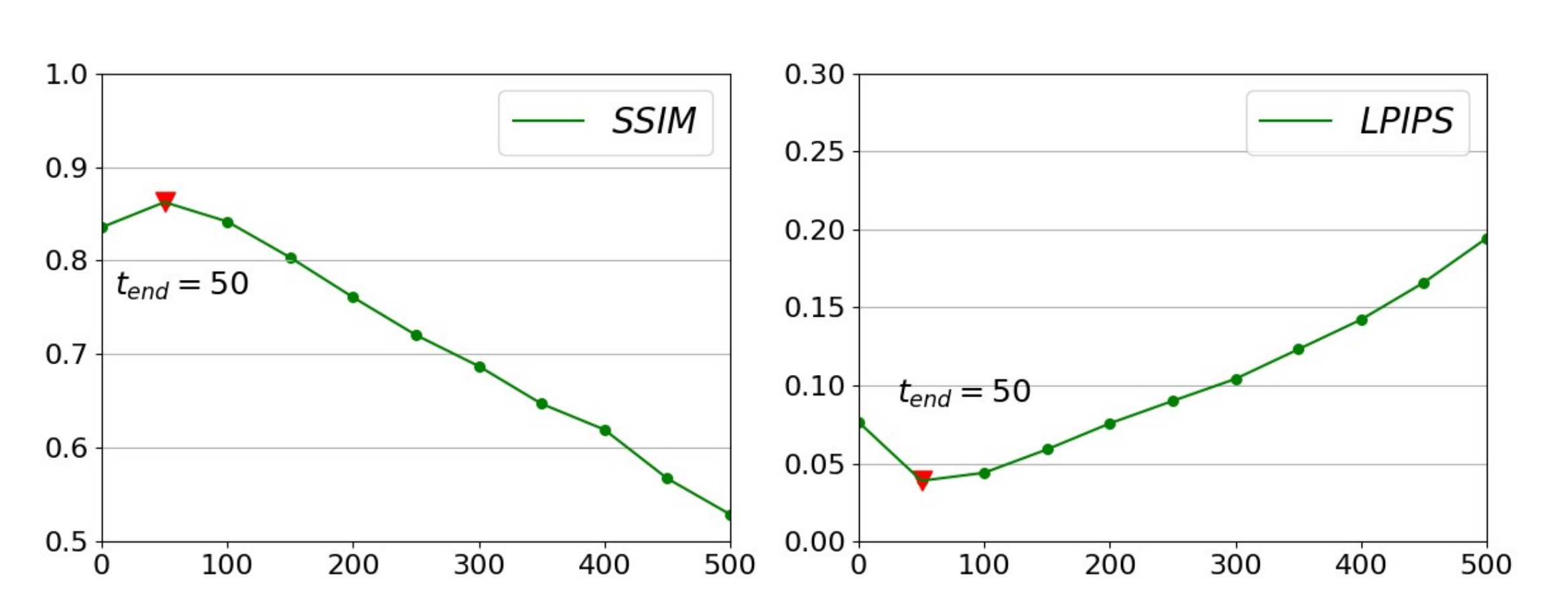}
    \caption{Quantitative evaluation at different $t_{end}$. The left shows the SSIM results and the right presents the LPIPS results of $t_{end}$, both reaching their optimum at $t_{end}=50$.}
    \label{Fig:LRS}
\end{figure}

In the initialization stage of TD-BFR, a good initial condition helps to facilitate more efficient restoration in the subsequent stages. In the first stage of TD-BFR, analysis from ~\cite{dr2} indicates that with $N^0=2$, semantics can be preserved to the fullest extent although substantial removal of image degradation is not achieved, which means samples from smaller $N^0$ are more similar to the input but still contain degradation. Therefore, we utilize $N^0=2$ at the resolution of 64 as the starting parameter. Consequently, we fix $N^0=2$ and set the length of $t^0$ to 50 steps, which means $t_{begin} - t_{end} = 50$ (the entire steps extend to 1000 steps). To provide the most optimal startup, we select SSIM and LPIPS on Various $t_{end}$ as references for evaluation. As depicted in \cref{Fig:LRS}, the findings indicate that the optimal SSIM and LPIPS scores are achieved when $t_{end}=50$. Therefore, we adopt $t_{end}=50$ as the initialization parameter for LRS.

\section{Experiments}

\subsection{Implementation, Datasets, Metrics}
\noindent \textbf{Implementation.} 
The complete diffusion model at 64 resolution employed by LRS is trained on the FFHQ dataset~\cite{GFPGAN35}, comprising 70,000 high-quality facial images. Subsequently, we utilize the pretrained DDPM proposed by \cite{ILVR} for the second stage. Following this, we utilize Stable Diffusion 2.1-base~\cite{ldm} as the generative prior and fine-tune in the way of concatenating extra channels for 87.5k iterations (with a batch size of 8) on the degraded FFHQ dataset~\cite{GFPGAN35}. The specific parameters for degradation are detailed in the appendix.
The optimization process employs Adam \cite{adam} as the optimizer, with a fixed learning rate of $5\times10^{-5}$ throughout all iterations. Training is performed on images of resolution $512\times512$ using 1 NVIDIA A100 GPU.

\noindent \textbf{Specific parameters for degradation of training datasets.}
The degraded FFHQ is outlined below:
\begin{equation}
    \vspace{-0.5mm}
    y = [(x \circledast k_\sigma) \downarrow_r + n_\delta]_{JPEG_q}.
    \label{Eq:deg}
\end{equation}
Initially, there is a 50\% probability that the image undergoes convolution with a blur kernel randomly selected from Gaussian blur, mean blur, and median blur, each with kernel sizes ranging from 3 to 15. Additionally, motion blur, with kernel sizes ranging from 5 to 25, is applied with the same probability. Subsequently, the blurred image is downsampled to a resolution of 64 using a scale factor of $r=8$. Gaussian noise, in the range of $\{0, 0.1^*255\}$, is then added to the image with a probability of 20\%. Finally, JPEG compression is performed on the image with a probability of 70\%, using compression qualities ranging from 10 to 65.

\noindent \textbf{Specific parameters for Test Datasets}. We construct one synthetic dataset on 
CelebA-HQ~\cite{GFPGAN51} and three real-world datasets, including WIDER-Test~\cite{CF41}, CelebChild, and LFW-Test~\cite{GFPGAN29}  for testing. For the synthetic dataset, which is a high-quality version of CelebA that consists of 30,000 images at 1024×1024 resolution, we utilize a degradation model commonly employed in previous studies ~\cite{SPAR}. Specifically, we apply the degradation model mentioned before to synthesize testing data from CelebA-HQ. In this degradation model, the high-quality image $x$ undergoes convolution with a Gaussian blur kernel, with a 50\% probability, varying in size from 3 to 9. Subsequently, it is downsampled to a resolution of 64, employing a scale factor of $r=8$. Gaussian noise within the range of 5 to 30 and Poisson noise within the range of 0 to 10 are then introduced with a probability of 0.5. Finally, JPEG compression is applied with a probability of 0.5, utilizing compression quality levels ranging from 30 to 95. 

For real-world datasets, WIDER-Test comprises 400 regular cases from the WIDER-face dataset, CelebChild includes 180 images of child celebrities sourced from the internet, and LFW-Test consists of 1711 LQ images with unknown degradation obtained from the internet.

\noindent \textbf{Metrics.} 
For evaluation, we adopt two pixel-wise metrics (PSNR and SSIM), a reference perceptual metric (LPIPS~\cite{lpips}), and two non-reference perceptual metrics (NIQE~\cite{niqe} and FID ~\cite{GFPGAN27}). All metrics are used for the evaluation of synthetic data while only NIQE and FID are used for real-world datasets due to the lack of referential HQ images.

\section{Complete qualitative experiment on real-world datasets}
\begin{figure*}
    \centering\vspace{-5mm}
    \includegraphics[width=1.0\textwidth]{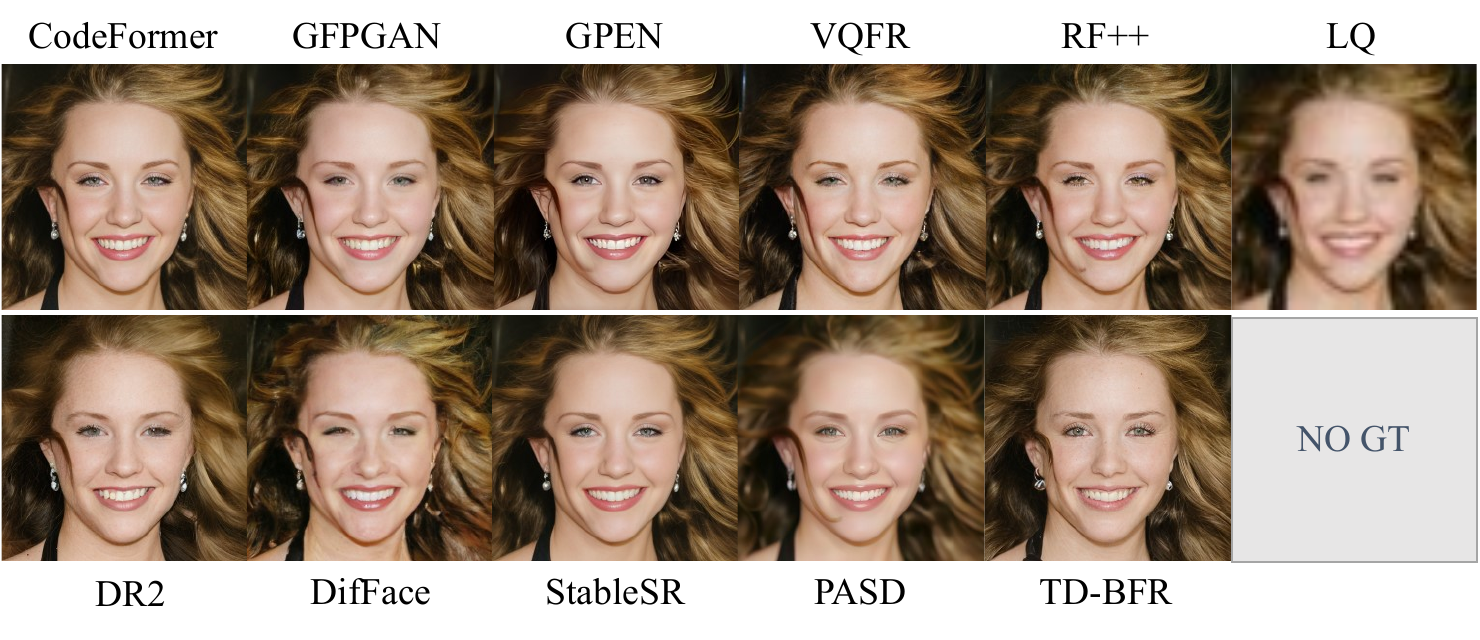}
    \vspace{-5mm}
    \caption{Complete qualitative results on LFW dataset. (Zoom in for the best view)}
    \label{fig:complete_LFW}
    \vspace{-5mm}
\end{figure*}

\begin{figure*}
    \centering
    \vspace{-5mm}
    \includegraphics[width=1.0\textwidth]{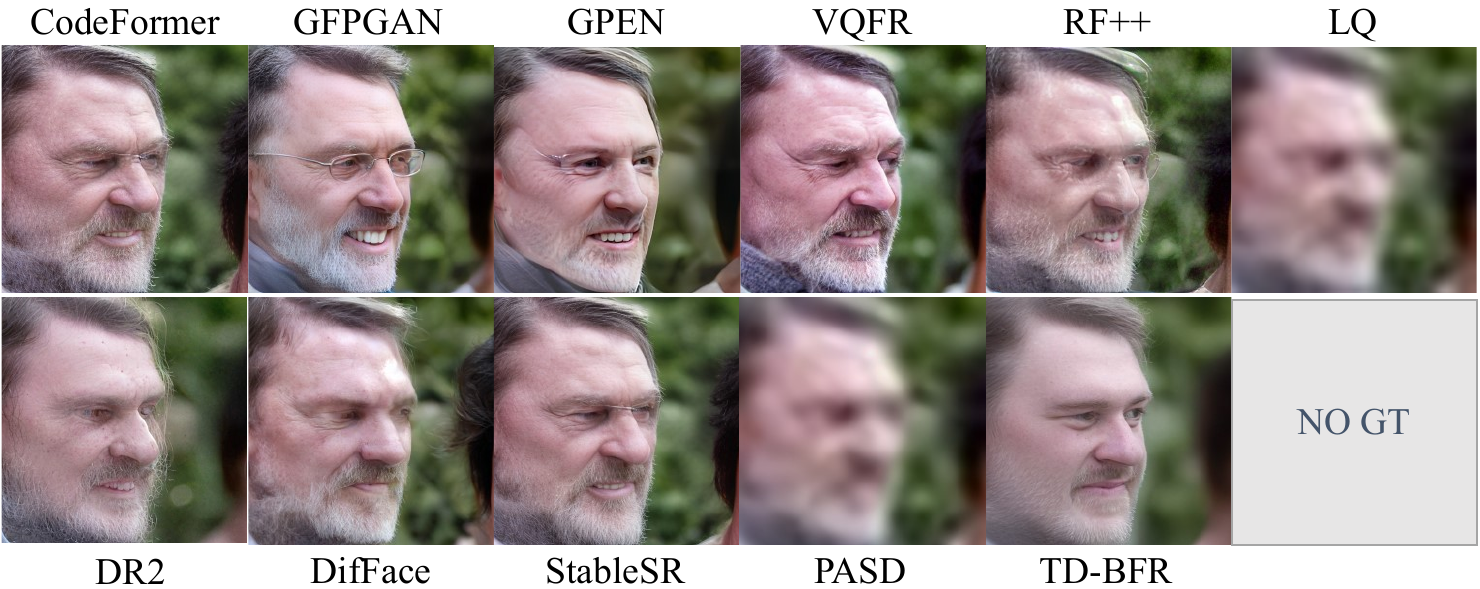}
    \vspace{-5mm}
    \caption{Complete qualitative results on Wider dataset. (Zoom in for the best view)}
    \vspace{-10mm}
    \label{fig:complete_Wider}
    
\end{figure*}

Due to space constraints in the manuscript, not all comparative experimental results are fully presented. Therefore, the complete qualitative results are shown in the supplementary, as illustrated in \cref{fig:complete_child}, \cref{fig:complete_LFW}, \cref{fig:complete_Wider}.

\begin{figure*}
    \centering
    \vspace{-10mm}
    \includegraphics[width=1.0\textwidth]{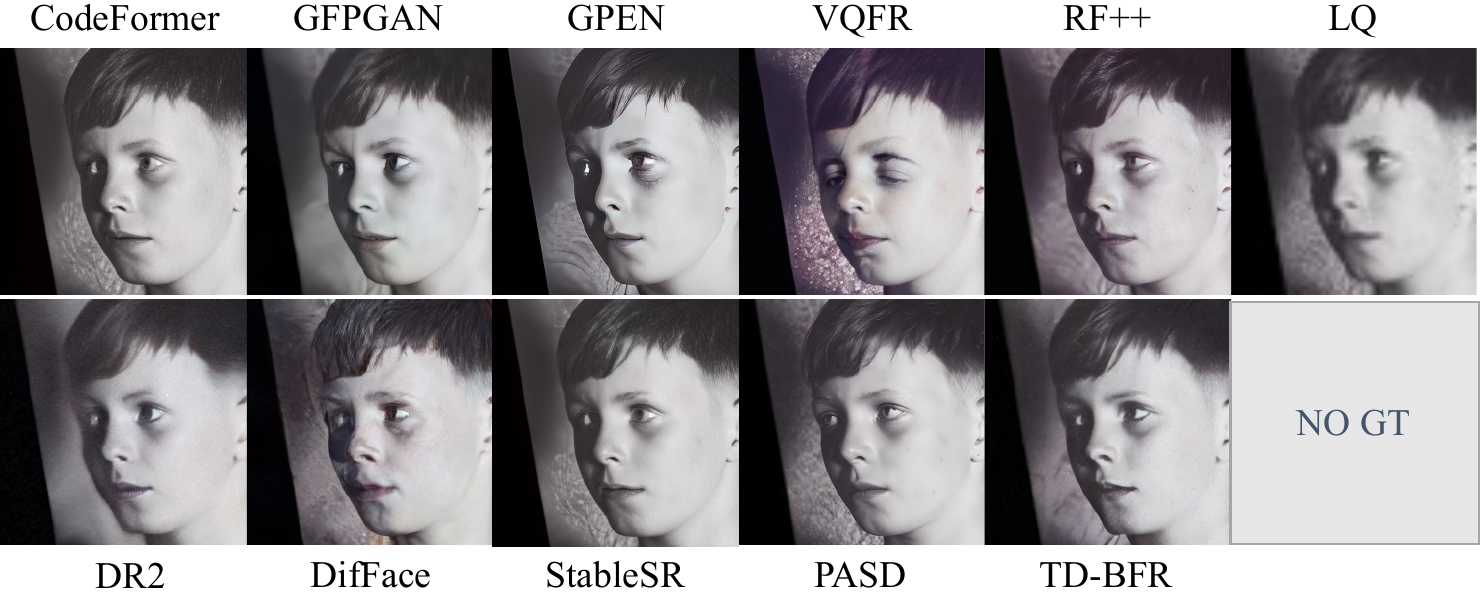}
    \vspace{-5mm}
    \caption{Complete qualitative results on Celeb-Child dataset. (Zoom in for the best view)}
    \label{fig:complete_child}
    \vspace{-5mm}
\end{figure*}

\section{Additional Results on Blind Face Restoration Task}
\begin{figure*}
    \centering
    \includegraphics[width=1.0\textwidth]
    {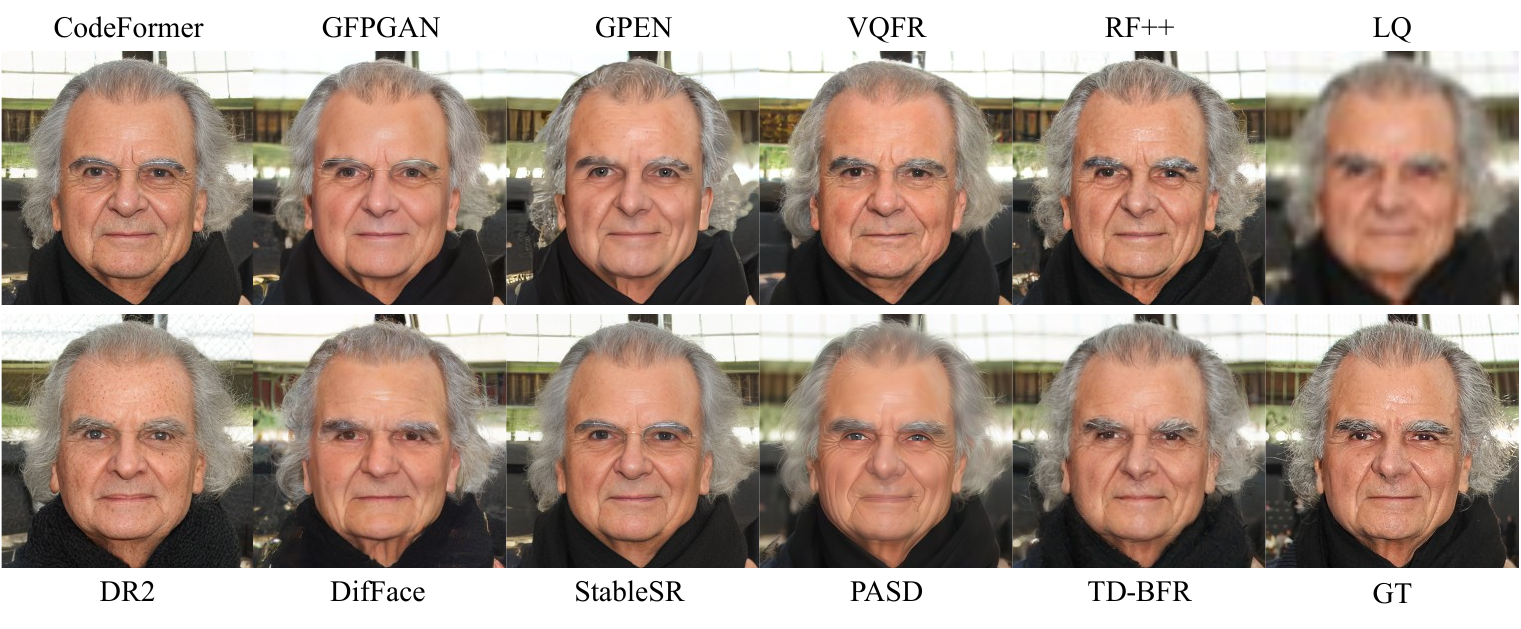}
    \caption{Additional results on both synthesis and real datasets. (Zoom in for the best view)}
    
    \label{Fig:additional_result}
\end{figure*}
We additionally show some blind face restoration results in the appendix as shown in \cref{Fig:additional_result}. Qualitative results show that our TDBFR method can significantly save sampling time while still maintaining superior restoration effects.

\bibliographystyle{IEEEbib}
\bibliography{icme2025references}